\documentclass[10pt,twocolumn,letterpaper]{article}

\usepackage{wacv}
\usepackage{times}
\usepackage{epsfig}
\usepackage{graphicx}
\usepackage{amsmath}
\usepackage{amssymb}

\usepackage{subcaption}

\usepackage{pdfpages}

\usepackage{url}
\usepackage{algorithm2e}

\usepackage{soul} 

\usepackage{array}
\newcolumntype{C}[1]{>{\centering\let\newline\\\arraybackslash\hspace{0pt}}m{#1}}

\wacvfinalcopy 

\begin{document}
\title{Summarization of ICU Patient Motion from Multimodal Multiview Videos}

\author{Carlos Torres\textsuperscript{\dag} ~~~ Kenneth Rose\textsuperscript{\dag} ~~~ Jeffrey C. Fried\textsuperscript{*} ~~~ B.S. Manjunath\textsuperscript{\dag} \\
\textsuperscript{\dag}University of California Santa Barbara ~~ \textsuperscript{*}Santa Barbara Cottage Hospital \\
{\tt\footnotesize{ \{carlostorres, rose, manj\}@ece.ucsb.edu ~ jfried@sbch.org} }
}

\maketitle

\begin{abstract}
Clinical observations indicate that during critical care at the hospitals, patients sleep positioning and motion affect recovery. Unfortunately, there is no formal medical protocol to record, quantify, and analyze patient motion. There is a small number of clinical studies, which use manual analysis of sleep poses and motion recordings to support medical benefits of patient positioning and motion monitoring. Manual processes are not scalable, are prone to human errors, and strain an already taxed healthcare workforce. This study introduces DECU (Deep Eye-CU): an autonomous mulitmodal multiview system, which addresses these issues by autonomously monitoring healthcare environments and enabling the recording and analysis of patient sleep poses and motion. DECU uses three RGB-D cameras to monitor patient motion in a medical Intensive Care Unit (ICU). The algorithms in DECU estimate pose direction at different temporal resolutions and use keyframes to efficiently represent pose transition dynamics. DECU combines deep features computed from the data with a modified version of Hidden Markov Model to more flexibly model sleep pose duration, analyze pose patterns, and summarize patient motion. Extensive experimental results are presented. The performance of DECU is evaluated in ideal (BC: Bright and Clear/occlusion-free) and natural (DO: Dark and Occluded) scenarios at two motion resolutions in a mock-up and a real ICU. The results indicate that deep features allow DECU to match the classification performance of engineered features in BC scenes and increase the accuracy by up to $8\%$ in DO scenes. In addition, the overall pose history summarization tracing accuracy shows an average detection rate of $85\%$ in BC and of $76\%$ in DO scenes. The proposed keyframe estimation algorithm allows DECU to reach an average $78\%$ transition classification accuracy.
\end{abstract}

\paragraph*{Keywords:}
\noindent Healthcare, Multimodal, Multiview, Deep Features, Hidden Markov Models, Multimodal Emission, Pose Transitions, ICU Monitoring, Motion Summarization.
\section{Introduction}\label{sec:intro}
The recovery rates of patients admitted to the ICU with similar conditions vary vastly and often inexplicably~\cite{giraud1993iatrogenic}. ICU patients spend most of their time on a bed, while cycling over various decubitus positions. The rate and range of patient motion are believed to be indicators of distress and increased or decreased recovery~\cite{von1984consequences}. Although patients are continuously monitored by staff, there are no procedures to reliably analyze and understand pose variations from patient observations such as videos. Nevertheless, limited clinical studies~\cite{morris2007moving} suggest that patient health-dependent positioning and controlled motion enhance patient recovery, while inadequate poses and uncontrolled or erratic motion aggravates wounds and injuries. 

While recording and analyzing the  motion of patients using human observers is a straightforward solution, it puts strain on an already taxed healthcare workforce. It does not scale with the volume of the data and is prone to human errors. This work introduces DECU, a multimodal multiview autonomous system for patient positioning and motion analysis. DECU enables the following analytical features for healthcare: 
\begin{enumerate}
    \item motion quantification (rate and range) to aid the analysis and prevention of decubitus ulcers (DUs) or bed sores;
    \item timely detection of erratic, harmful, or distressed motion that can be used to stop patients from pulling intra-venous lines or falling off the bed; and
    \item summarization of pose sequences (pose history) over extended periods of time, which can be used to evaluate quality sleep quality without intrusive equipment.
\end{enumerate}

\begin{figure*}[!h]
    \begin{center}
     	\includegraphics[width=1\linewidth]{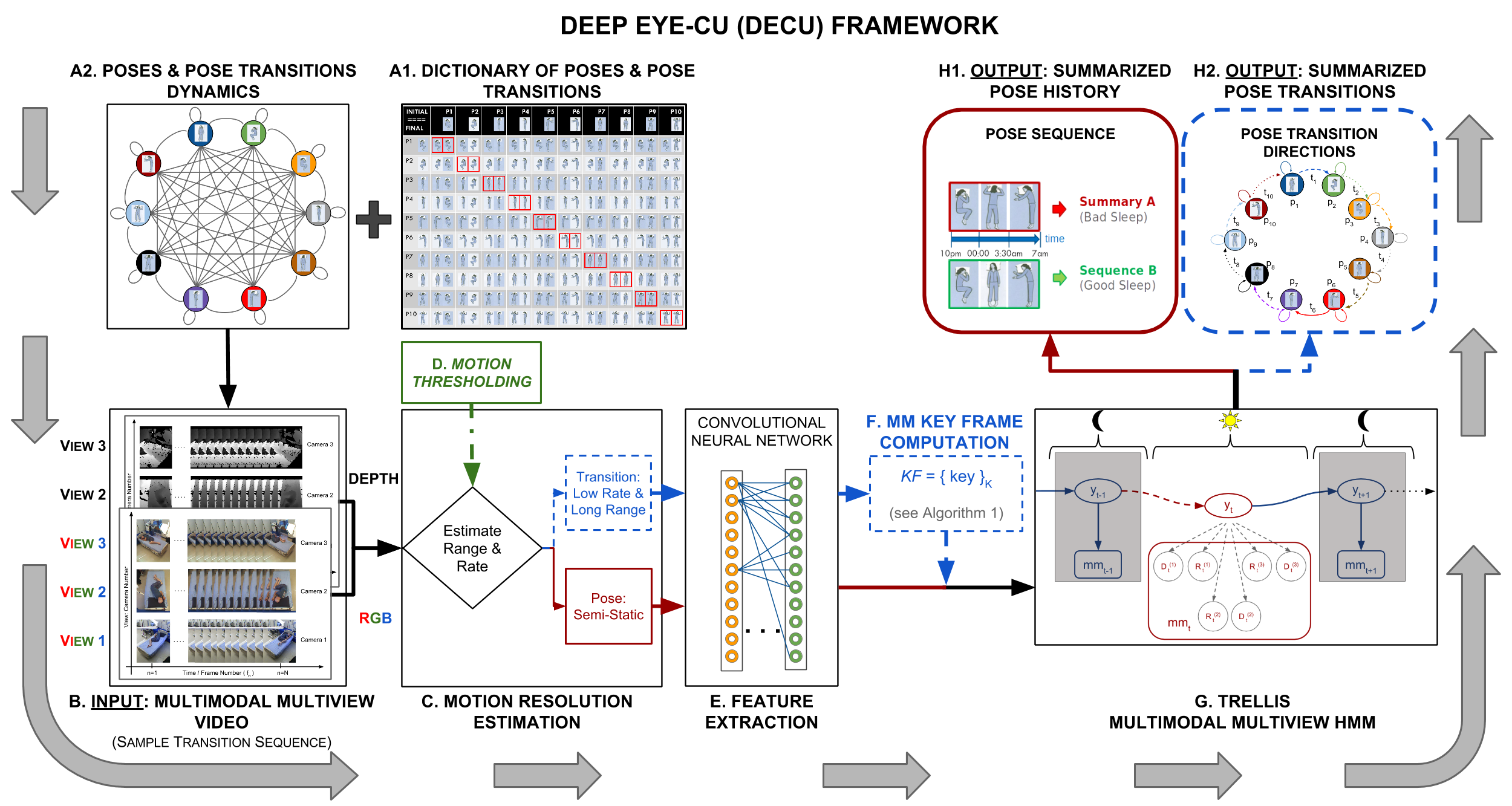}
    \end{center}
    \vspace{-.5cm}
    \caption{Stages of the DECU framework, which uses multimodal multiview ($MM$) data and the modified Hidden Semi-Markov Modeling to monitor patient motion. From left-to-right (A to H): the set of references is shown on stage A (top-left); (A1) a dictionary of poses and pose transitions, and (A2) a lattice showing possible motion dynamics between poses. Stage B (bottom-left) shows the multimodal multiview input video. Stage C (center-left) selects the summarization resolution and activates keyframe identification when required. Stage D (center) integrates the motion thresholds (estimated at training) to account for various levels of motion resolution and sensor noise. Stage E (bottom-center) represents the feature extraction block via a convolutional neural network. Stage F (center-right) shows the keyframe identification process using algorithm \ref{algo:keyframes}. Stage G (bottom-right) shows the multimodal multiview HMM trellis, which encodes illumination and occlusion variations. Stage H (top-right) shows the two possible summarization outputs (H1) pose history and (H2) pose transitions.}
    \label{fig:overview}
\end{figure*}

DECU incorporates algorithms for keyframe extraction and pose (state) duration estimation to autonomously and unobtrusively monitor patient motion at different temporal resolutions. DECU combines deep features from multimodal multiview data with Hidden Semi-Markov Models (HSMM) to more flexible model pose durations. DECU extracts keyframes from multiple sources to reliably represent transitions and monitor and summarize patient motion. DECU is designed, trained, and tested in a mock-up ICU and tested in a real ICU. Fig \ref{fig:overview} shows the major elements of the framework (stages A-H). Stage A (top right) contains the references. Stage B (bottom left) shows frames from a sample sequence recorded using multimodal (RGB and Depth) multiview (three cameras) sources. At stage C, the framework selects the summarization resolution and activates the keyframe identification stage (for training). Stage D contains the motion thresholds (dense optic-flow estimated at training) to distinguish between the motion types and account for depth sensor noise. Deep features are extracted at stage E. Stage F shows the keyframe computation, which compresses motion and encodes motion segments (encoding of duration of poses and transitions). Stage G shows the multimodal multiview HMM trellis with two scene conditions. Finally, stage H shows the results: pose history and pose transition summarizations.

DECU summarization is evaluated in ideal (BC: Bright and Clear/occlusion-free) and natural (DO: Dark and Occluded) scenarios at two motion resolutions in a mock-up and a real ICU. Experimental results indicate that using deep features for pose representation allows DECU to match the classification performance of engineered features in BC scenes and increases the accuracy by up to 8\% in DO  scenes. The overall pose history summarization (coarser time resolution) tracing accuracy shows an average detection rate of 85\% in BC scenes and of 76\% in DO scenes. The performance of pose transition summarization (finer time resolution) depends directly on the range of motion, dissimilarity between poses, and direction of rotation. The proposed multimodal multiview keyframe estimation algorithm allows DECU to reach a mean transition classification accuracy of 78\% using  a maximum of five pseudo-poses (keyframes) to represent a transition.

\subsection{Background}\label{subsec:backgnd}
In August 2016, Harvard Medical School published a report stating that monitoring ICUs can save up to \$15 billion by saving \$20,000 in each of the 750,000 ICU beds in the U.S. with preventive care and by reducing the effect of preventable ICU-related conditions such as poor quality of sleep and DUs. The U.S. Department of Health and Human Services \footnote{U.S. Department of Health Services -- online report Feb 2016} states that the U.S. ICU expenditure is about \$130 million per year and, at its current state, rises by \$5 billion per year. The ICUs in the U.S. receive about five million patients per year. The average ICU stay is 9.3 days and patient mortality rate ranges from 10 to 30\% depending on health conditions.

Clinical studies covering sleep analysis indicate that sleep hygiene directly impacts healthcare. In addition, quality of sleep and effective patient rest are correlated to shorter hospital stays, increased recovery rates, and decreased mortality rates. Clinical applications that correlate body pose and movement to medical conditions include sleep apnea -- where the obstructions of the airway are affected by supine positions \cite{sahlin2009sleep}. Pregnant women are recommended to sleep on their sides to improve fetal blood flow \cite{morong2015sleep}. The findings of \cite{bihari2012factors}, \cite{idzikowski2003sleep}, and \cite{weinhouse2006sleep} correlate sleep positions with quality of sleep and its various effects on patient health. Decubitus ulcers (bed sores) appear on bony areas of the body and are caused by continuous decubitus positions \footnote{Online Medical Dictionary}. Although nefarious, bed sores can be prevented by manipulating patient poses over time. Standards of care require that patients be rotated every two hours. However, this protocol has very low compliance and in the U.S., a very high number of ICU patients develop DUs \cite{soban2011preventing}. There is little understanding about the set of poses and pose durations that cause or prevent DU incidence. Studies that analyze pose durations, rotation frequency, rotation range, and the duration of weight/pressure off-loading are required, as are the non-obtrusive measuring tools to collect and analyze the relevant data. Additional studies analyze pose manipulation effects on treatment of severe acute respiratory failure such as: ARDS (Adult Respiratory Distress Syndrome), pneumonia, and hemodynamics in patients with various forms of shock.  These examples highlight the importance of DECU's autonomous patient monitoring and summarization tasks. They accentuate the need and challenges faced by the framework, which must be capable of adapting to hospital environments and supporting existing infrastructure and standards of care.

\subsection{Related Work}\label{subsec:related}
There is a large body of research that focuses on recognizing and tracking human motion. The latest developments in deep features and convolutional neural network architectures achieve impressive performance; however, these require large amounts of data \cite{cheron2015p}, \cite{veeriah2015differential}, \cite{baccouche2011sequential}, and \cite{tran2015learning}. These methods tackle the recognition of actions performed at the center of the camera plane, except for \cite{soran2015generating}, which uses static cameras to analyze actions. Method \cite{soran2015generating} allows actions to not be centered on the plane; however, it requires scenes with good illumination and no occlusions. At its current stage of development the DECU framework cannot collect the large number of samples necessary to train a deep network without disrupting the hospital.

Multi-sensor and multi-camera systems and methods have been applied to smart environments \cite{hoque2012aalo} and \cite{wu2010multiview}. The systems require alterations to existing infrastructure making their deployment in a hospital logistically impossible. The methods are not designed to account for illumination variations and occlusions and do not account for non-sequential, subtle motion. Therefore, these systems and methods cannot be used to analyze patient motion in a real ICU where patients have limited or constrained mobility and the scenes have random occlusions and unpredictable illumination. 

Healthcare applications of pose monitoring include the detection and classification of sleep poses in controlled environments \cite{huang2010multimodal}. Static pose classification in a range of simulated healthcare environments is addressed in \cite{torres2015multimodal}, where the authors use modality trust and RGB, Depth, and Pressure data. In \cite{torres2016EyeCU}, the authors introduce a coupled-constrained optimization technique that allows them to remove the pressure sensor and increase pose classification performance. However, neither method analyzes poses over time or pose transition dynamics. A pose detection and tracking system for rehabilitation is proposed in \cite{obdrvzalek2012real}. The system is developed and tested in ideal scenarios and cannot be used to detect constrained motion. In \cite{padoy2009workflow} a controlled study focuses on work flow analysis by observing surgeons in a mock-up operating room. A single depth camera and Radio Frequency Identification Devices (RFIDs) are used in \cite{lea20133d} to analyze work flows in a Neo-Natal ICU (NICU) environment. These studies focus on staff actions and disregard patient motion. Literature search indicates that the DECU framework is the first of its kind. It studies patient motion in a mock-up and a real ICU environment. DECU's technical innovation is motivated by the shortcomings of previous studies. It observes the environment from multiple views and modalities, integrates temporal information, and accounts for challenging natural scenes and subtle patient movements using principled statistics. 
\subsection{Proposed Approach}\label{subsec:proposed}
DECU is a new framework to monitor patient motion in ICU environments at two motion resolutions. Its elements include time-series analysis algorithms and a multimodal multiview data collection system. The algorithms analyze poses at two motion resolutions (sequence of poses and pose transition directions). The system is  capable of collecting and representing poses from multiview multimodal data. The views and modalities are shown in Figure \ref{fig:ICU} (a) and (b). A sample motion summary is shown in Figure \ref{fig:ICU} (c). Patients in the ICU are often bed-ridden or immobilized. Overall, their motion can be unpredictable, heavily constrained, slow and subtle, or aided by caretakers. The two resolutions address different medical needs. Pose history summarization is the coarser resolution. It provides a pictorial representation of poses over time (i.e., the history). The applications of the pose history include prevention and analysis of DUs and analysis of sleep-pose effects on quality of sleep. The pose transition summarization is the finer resolution. DECU looks at the transition poses that occur while a patient transitions between two clearly defined sleep poses. Physical therapy evaluation is one application of transition summarization. 

\paragraph{Main Contributions of this work:}
\begin{enumerate}
    \item An adaptive framework capable of monitoring patient motion at various resolutions. The algorithm detects patient motion behavior and summarizes the sequence of sleep poses and the subtle motion and direction between two poses using segments.
    
    \item A non-disruptive and non-obtrusive monitoring system robust to natural healthcare scenarios and conditions such as variable illumination and partial occlusions.
    
    \item An algorithm that effectively compresses sleep pose transitions using a subset of the most informative and most discriminative keyframes. The algorithm incorporates data from all views and modalities to identify keyframes and increase monitoring resolution.

    \item A fusion technique that incorporates observations from multiple modalities and views into emission probabilities to leverage complementary information and estimate intermediate poses and transitions over time.
\end{enumerate}

\section{DECU System Description} \label{sec:System}
The DECU system is modular and adaptive. It is composed of three nodes and each node has three modalities (RGB, Depth, and Mask). At the heart of each node is a Raspberry Pi3 running Linux Ubuntu, which controls a Carmine RGB-D cameras\footnote{Primesense, manufacturer of Carmine sensors, was acquired by Apple Inc. in 2013; however, similar devices can be purchased from structure.io}. The units are synchronized using TCP/IP communication. DECU combines information from multiple views and modalities to overcome scene occlusions and illumination changes.
\subsection{Multiple modalities (Multimodal).} Multimodal studies use complementary modalities to classify static sleep poses in natural ICU scenes with large variations in illumination and occlusions. This study uses the findings from those studies (which are also validated in the experimental section) regarding the benefits of multimodal systems.

\paragraph{RGB (R)} Standard RGB video data provides reliable information to represent and classify human sleep poses in scenes with relatively ideal conditions. However, most people sleep in imperfectly illuminated scenarios, using sheets, blankets, and pillows that block and disturb sensor measurements. The systems collects RGB color images of dimensions $640\times480$ from each actor in each of the scene conditions, and extracts pose appearance features representative of the lines in the human body (i.e., limbs and extremities).

\paragraph{Depth (D)} Infrared depth cameras can be resilient to illumination changes. The Eye-CU system uses Primense Carmine devices to collect depth data. The devices are designed for indoor use and can acquire images of dimensions $640\times480$. These sensors use 16 bits to represent pixel intensity values, which correspond to the distance from sensor to a point in the scene. Their operating distance range is $0.8$ m to $3.5$ m; and their spatial resolution for scenes $2.0$ m away is $3.5$ mm for the horizontal (x) and vertical (y) axes, and $30$ mm along the depth (z) axis. The systems uses the depth images to represent the 3-dimensional shape of the poses. The usability of these images, however, depends on depth contrast, which is affected by the deformation properties of the mattress and blanket present in ICU environments.

\subsection{Multiple views (Multiview).} The studies from \cite{torres2016EyeCU} and \cite{ramagiri2011real} show that analyzing actions from multiple views and multiple orientations greatly improves detection. In particular, the studies indicate that multiple views provide algorithmic view and orientation Independence.
    
\subsection{Time Analysis (Hidden Semi-Markov Models).} ICU patients are often immobilized or recovering. They move subtly and slowly (very different from the walking or running motion). DECU effectively monitors subtle and abrupt patient motion by breaking the motion cues into segments.

\section{Data Collection}\label{sec:Data}
\begin{figure*}[t]
    \begin{center}
    \begin{tabular}{ccc}
        \fbox{\includegraphics[width=5.5cm]{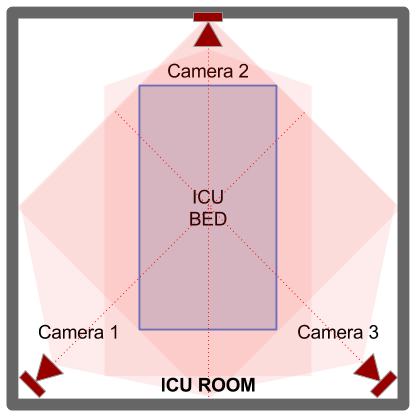}}              &
        \fbox{\includegraphics[width=5.575cm]{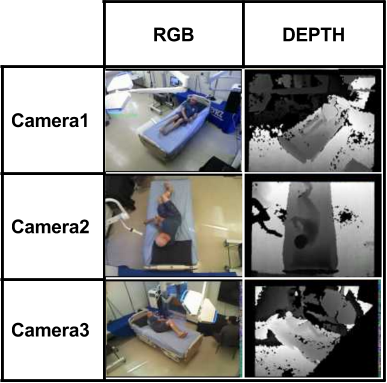}}             &
        \fbox{\includegraphics[width=4.6cm]{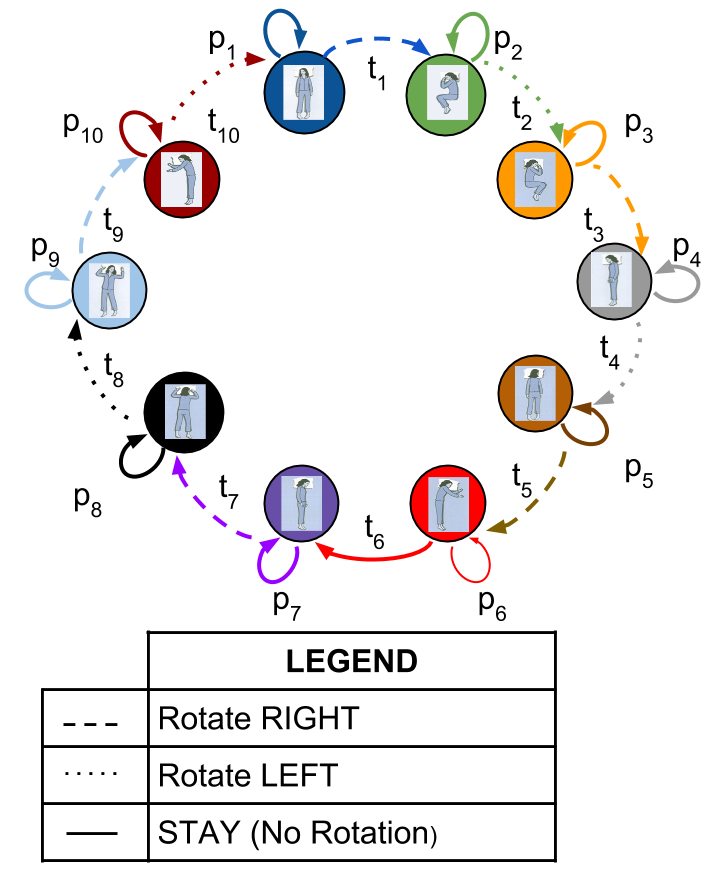}} 

         \\
        (a)&(b)&(c)
    \end{tabular}
    \end{center}
    \vspace{-.5cm}
    \caption{The transition data is collected in a mock-up ICU and a real ICU: (a) shows the relative position of the cameras with respect to the ICU room and ICU bed; (b) shows a set of randomly selected poses and pose transitions, which are represented by lines (dashed, dotted, and solid lines defined in the legend box); (c) shows a sample set of sleep-pose transitions and rotation directions.}
    \label{fig:ICU}
\end{figure*}

Pose data is collected in a mock-up ICU with seven actors and tested in medical ICU with two real patients. The diagram in Figure \ref{fig:ICU} (b) shows the top-view of the rigged mock-up ICU room and the camera views. In the mock-up ICU, actors are asked follow the same test sequence of poses. The sequence is set at random using a random number generator. Figure \ref{fig:ICU} (c) shows a sequence of $20$ observations, which include ten poses ($p_1$ to $p_{10}$) and ten transitions ($t_1$ to $t_{10}$) with random transition direction.

All actors in the mock-up ICU are asked to assume and hold each of the poses while data is being recorded from multiple modalities and views. A total of $28$ sessions are recorded: $14$ under ideal conditions (BC: bright and clear) and $14$ under challenging conditions (DO: dark and occluded). The annotated dataset will available at \textcolor{blue}{\url{vision.ece.ucsb}}.

\subsection{ICU Rooms Infrastructure.}
The DECU system and algorithmic elements are designed, tested, and refined in a mock-up ICU with actors and simulated hospital scenarios. Once ready for real-world testing, DECU is deployed in a medical ICU with real patients and hospital scene conditions, and where medical experts evaluate its benefits. 

\paragraph{The mock-up ICU room}
The mock-up ICU room allows researchers to collect data, design and test algorithms, and evaluate and refine the DECU system and algorithms. Three views of the mock-up ICU are shown in Figure \ref{fig:ICU}.

\begin{figure*}[h!]
    \begin{center}
        \includegraphics[width=1\linewidth]{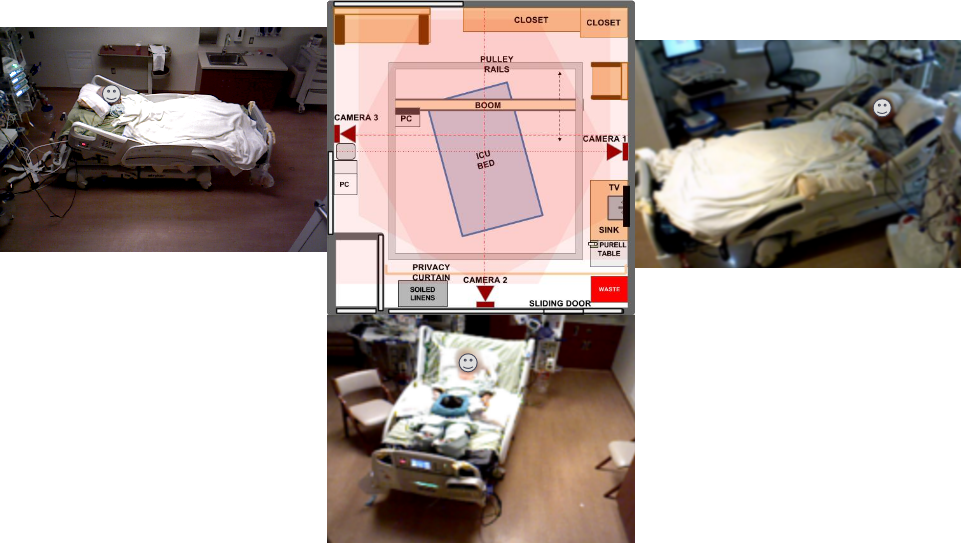}
    \end{center}
    \vspace{-.5cm}
    \caption{Top view of the node locations (center of the image) and views of the real a medical ICU room and ICU patient.}
    \label{fig:medical_ICU}
\end{figure*}

\paragraph{The real ICU room}
DECU is currently deployed in a real ICU at a local community hospital where medical experts validate its benefits and performance and explore its applications. The system nodes are battery powered and the three nodes account for unexpected occlusions and illumination changes. Views of the medical ICU are shown in Figure \ref{fig:medical_ICU}

\subsection{Pose Transitions.}
The actors follow the sequence poses and transitions shown in Stage A from Figure \ref{fig:overview}. Each initial pose has 10 possible final poses (inclusive) and each final pose can be arrived to by rotating left or right. The combination of pose pairs and transition directions generates a set of 20 sequences for each initial pose. There are 10 possible initial poses. One actor and one recording session generates 200 sequence pairs.

\subsection{Feature Selection.}
Previous findings indicate that engineered features such as geometric moments (gMOMs) and histograms of oriented gradients (HOG) are suitable for the classification of sleep poses. However, these features are limited in their ability to represent body configurations in dark and occluded scenarios. The latest developments in deep learning and feature extraction led this study to consider deep features extracted from the VGG \cite{simonyan2014very}  and the Inception \cite{szegedy2015going} architectures. Experimental results (see Sec \ref{sec:exp_FeatValidation}) indicate that Inception features perform better than gMOMs, HOG, and VGG features. Parameters for gMOM and HOG extraction are obtained from \cite{torres2015multimodal}. Background subtraction and calibration procedures from \cite{Hartley2004} are applied prior to feature extraction. 

\section{Problem Description}
Patients in the ICU spend most of their time in bed and their motion is limited to a small set of poses. Practitioners manipulate the patient poses to prevent DUs, evaluate sleep hygiene, and enhance recovery rates among other. DECU uses videos from multiple views and modalities to monitor patient poses and their transitions. However, it is necessary that the system and the algorithms properly handle motion rates (speed) and motion ranges. For instance, pose history summarization analysis patients poses over a longer period of time (e.g., eight hours) at a very low sampling rate. The pose transition summarization is another example. The summarization and the analysis involves identifying the set of pseudo poses associated with the transition and quantifying the direction of rotation. The are various challenges in DECU. The first main challenge arouses by conventional algorithms being unable model pose durations effectively. The second challenges involves detecting the direction of motion and rotation when transitioning between poses. The last main challenge involves accurate representation of pseudo poses and keyframe estimation. The challenges and approaches are discussed in this section.

The DECU system uses $M$ multimodal cameras stationed at different locations to obtaining $V$ views of the patients and estimate pose transitions such as the one shown in Figure~\ref{fig:ICU} (c). Note that there are two directions of rotation for the a patient or actor to transition from the faller facing up (\textit{falU}) position to the fetal laying left (\textit{fetL}) position. Features extracted from video frames $\mathcal{F} = \{f_t\}$, for $1 \leq t \leq T$ to construct feature vectors  $\mathbf{X} = X_{1:T}$ are used to represent non-directly observable poses ($\mathbf{Y} = Y_{1:T}$). The first objective of DECU is to find the sequence of poses ($\mathbf{Y} = Y_{1:T}$) that probabilistically can best represent the observations i.e., $\Pr\big(\mathbf{Y}, \mathbf{X}\big) = \Pr \big(Y_{1:T}, X_{1:T}\big)$.

Temporal patterns caused by sleep-pose transitions are simulated and analyzed using Hidden Semi-Markov Models (HSMMs) as described in Section \ref{sec:hsmm}. The interactions between the modalities to accurately represent a pose using different sensor measurements are encoded into the emission probabilities. Scene conditions are encoded into the set of states (the analysis of two scenes doubles the number of poses). Conventional Markov assumptions support DECU and ideally fit most of its analysis. However, HMMS are limited in their ability to distinguish between poses and pseudo-poses (i.e., transitory, short time body pose configurations observed when transitioning between poses) based on pose duration. This is because, by design, HMMs model the probability of staying in a given pose as a geometric distribution $\Pr_i(d) = {(a_{ii})}^{d-1} (1-a_{ii})$, where $d$ is the duration in pose $i$, and $a_{ii}$ is the self-transition probability of pose $i$. More details are discussed in this section and subsequent subsections. Table \ref{table:variables} describes the DECU variables.

\begin{table}[h!]
\scriptsize
\centering

\begin{tabular}{|l||l|}
\hline
\multicolumn{2}{c}{\bf{DECU VARIABLES}}                                           \\ \hline\hline
SYMBOL            & DESCRIPTION                                                   \\  \hline \hline
$\mathbf{A}$      & Transition probability matrix; $\mathbf{A} \in \mathbb{R} ^{|P| \times |P|}$ and $\mathbf{A} = \{a_{ij}\}$                                               \\
$a_{i,j}$         & Probability of transitioning from pose $i$ to $j$             \\
$\mathbf{B}$      & Emission probability matrix $\in \mathbb{R}^{|P|}$ and $\mathbf{B} = \{\mu_{in} \}$  \\
$b_u$             & Beginning of the $u$-th segment with $b_1 = 1$                \\
$D_k$             & $k$-th frame from the depth modality video                    \\
D                 & Face-Down patient pose                                        \\
$d$               & Segment duration                                              \\
$d_u$             & Segment duration for $u$-th segment                           \\
HMM               & Abbreviation for Hidden Markov Model                          \\
HSMM              & Abbreviation for Hidden Semi-Markov Model                     \\
$K$               & Data set size, $K = |\mathcal{X}|$                            \\ 
$k$               & Data point index, $1\leq k \leq K$                            \\
$KF$              & Set of keyframes representing a pose transition\\
L                 & Laying-Left patient pose                                      \\
$l$, $m$, and $n$ & Dummy variables                                               \\
$R_k$             & $k$-th frame from the rgb modality video                      \\
R                 & Laying-Right patient pose                                     \\

${\mu}_i$         & Probability that state $i$ generates the observation $x$ at time $t$      \\
$\boldsymbol{\pi}$    & Initial state probability vector $\in \mathbb{R}^ |P|$ and $\pi_i \in \boldsymbol{\pi}$  \\
$k$               & The time step index (i.e., $k=t$)                             \\
$P$               & Set of patient poses $P = \{p_i\}$                            \\
$P_{\text{mock}}$ & Set of actor poses in the mock-up ICU                         \\
$P_{\text{micu}}$ & Set of patient poses in the medical ICU (micu)                \\
$\Pr(Y,X)$        & Joint probability distribution between states and observations\\
$S$               & Set of time segments $S = \{s_u\}$ for $1 \leq u \leq U$      \\
$s$               & Segment element $s \in S$                                     \\ 
$t$               & Time tick with $1 \leq t \leq T$                              \\
$\tau_{td}$       & Store the estimated duration ($1\leq d \leq D$) at time ($t$) \\
$\theta$          & HMM model with probabilities $\mathbf{A}, \mathbf{B}$, and $\boldsymbol{\pi}$ \\

$U$               & Number of segments $U = |S|$ \\
U                 & Face-Up patient pose                         \\

$u$               & Segment index: $1 \leq u \leq U$    \\

$\mathcal{V}$     & View set $\mathcal{V}=\{\text{left,}\text{ center,} \text{ right} \}$ \\
$V$               & Number of views $V= |\mathcal{V}|$                          \\
$v$               & View index, $1\leq v \leq V$                                \\

$y_k$             & $k$-th hidden state $y_k \in \mathbf{Y}$                    \\
$\mathbf{Y}$      & Sequence of hidden states $|\mathbf{Y}| = T$                \\
$\mathcal{X}$     & Dataset indexed by $k$ (i.e., $\mathcal{X}_k$)              \\
$\mathcal{X}_k$   & $k$-th datapoint with $\{f_{N_m}\}_k = \{f_R, f_D, f_P\}_k$ \\

$x_k$             & $k$-th observation feature vector \\
$\mathbf{x}^{(v)}_{km}$& The $k$-th observable variable from view $v$ and modality $m$    \\ 

$\delta$          & Kroenecker delta function                                   \\
$\delta_t$        & The maximum probability duration                            \\

$\theta$          & Dummy variable used in inference                            \\
$\zeta $          & Stores the state label (for a pose) of the previous segment \\
$\phi  $          & Stores the best duration                                    \\
$\psi_t(i)  $     & Stores the label with the best duration for time $t$ and state $i$\\
\hline
\end{tabular}
\caption{DECU variable symbols and their descriptions.}
\label{table:variables}
\end{table}

\subsection{Hidden Markov Models (HMMs)} \label{sec:hmm}
HMMs are a generative approach that models the various poses (pose history) and pseudo-poses (pose transitions summarization) as states. The hidden variable or state at time step $k$ (i.e., $t=k$) is $y_k$ (state$_k$ or pose$_k$) and the observable or measurable variables ($x^{(v)}_{k,m}$, the vector of image features extracted from the $k$-th frame, the $m$-th modality, and the $v$-th view) at time $t=k$ is $x_k$ (i.e., $x_k = x^{(v)}_{k,m}= \{R_k, D_k, ... M_k \}$). The first order Markov assumption indicates that at time $t$, the hidden variable $y_t$, depends only on the previous hidden variable $y_{t-1}$. At time $t$ the observable variable $x_t$ depends on the hidden variable $y_t$. This information is used to compute $P(Y,X)$ via:

\begin{equation}\label{eqn:hmm}
\small
    P\big(Y_{1:T}, X_{1:T}\big) = P(y_1)\prod_{t=1}^{T}P\big(x_t | y_t\big)  \prod_{t=2}^{T}P\big(y_t|y_{t-1}\big)
\end{equation}

\noindent where $P(y_1)$ is the initial state probability distribution $(\mathbf{\pi})$. It represents the probability of sequence starting $(t=1)$ at pose$_i$ (state$_i$). $P\big(x_t | y_t\big)$ is the observation or emission probability distribution $(\mathbf{B})$ and represents the probability that at time $t$ pose$_i$ (state$_i$) can generate the observable mumtimodal multiview vector $x_t$. Finally, $P\big(y_t | y_{t-1}\big)$ is the transition probability distribution $(\mathbf{A})$ and represents the probability of going from pose$_i$ to pose$_o$ (state $i$ to $o$). The HMM parameters are $\mathbf{A} = \{a_{ij}\}$, $\mathbf{B} = \{\mu_{in}\}$, and $\boldsymbol{\pi} = \{\pi_i\}$, discussed below.

\paragraph{Initial State Probability Distribution $(\mathbf{\pi})$.}
The initial pose probabilities are obtained from \cite{idzikowski2003sleep} and adjusted to simulate the two scenes considered in this study. The scene initial state probabilities $\mathbf{\pi}$ is shown in Table \ref{table:InitialStateProbas}.

\begin{table*}[t]
{\small
\centering
\begin{tabular}{|l||c|c|c|l||c|l|}

\hline
\multicolumn{7}{c}{\bf{Initial State Probability:} $\mathbf{\pi} = \{\pi_i\}$}   \\ \hline\hline

Pose Name      & Acronym             & Symbol       & State - BC     & Probability   & State - DO   & Probability      \\ \hline
Soldier Up     & \textit{solU}       & $p_1$        & $s_1 $         & $~0.03$       & $s_{11} $    & $~0.02$          \\ 
Fetal Right    & \textit{fetR}       & $p_2$        & $s_2 $         & $~0.145$      & $~s_{12}$    & $~0.07$          \\
Fetal Left     & \textit{fetL}       & $p_3$        & $s_3 $         & $~0.145$      & $~s_{13}$    & $~0.07$          \\
Log Right      & \textit{logR}       & $p_4$        & $s_4 $         & $~0.05$       & $~s_{14}$    & $~0.03$          \\
Soldier Down   & \textit{solD}       & $p_5$        & $s_5 $         & $~0.02$       & $~s_{15}$    & $~0.01$          \\
Yearner Left   & \textit{yeaL}       & $p_6$        & $s_6 $         & $~0.04$       & $~s_{16}$    & $~0.02$          \\
Log Left       & \textit{logL}       & $p_7$        & $s_7 $         & $~0.05$       & $~s_{17}$    & $~0.03$          \\
Faller Down    & \textit{falD}       & $p_8$        & $s_8 $         & $~0.05$       & $~s_{18}$    & $~0.02$          \\
Faller Up      & \textit{falU}       & $p_9$        & $s_9 $         & $~0.05$       & $~s_{19}$    & $~0.03$          \\
Yearner Right  & \textit{yeaR}       & $p_{10}$     & $s_{10}$       & $~0.04$       & $~s_{20}$    & $~0.02$          \\
Other          & other               & $p_0$        & $s_{0}$        & $~0.036 $     & $ s_{0} $    & $~0.073$         \\
\hline
\end{tabular}

\caption{Initial probability for each of the 10 poses. Notice that poses facing Up have a higher probability than the poses that face Down, while Left and Right poses are equally probable. Please note that there is a category for poses not covered in this study identifiable by the label Other and the symbol $p_{11}$. Also, note that one pose can have two states based on the BC and DO scene conditions.}
\label{table:InitialStateProbas}
} 
\end{table*}

\paragraph{State Transition Probability Distribution $(\mathbf{A})$.}
The transition probabilities are estimated for one pose to the next one for Left (L) and Right (R) rotation directions as indicated in the results from Figs. \ref{fig:BC_results_transitions} and  \ref{fig:DO_results_transitions}.

\begin{figure*}[t]
    \begin{center}
     	\fbox{\includegraphics[width=.9\linewidth]{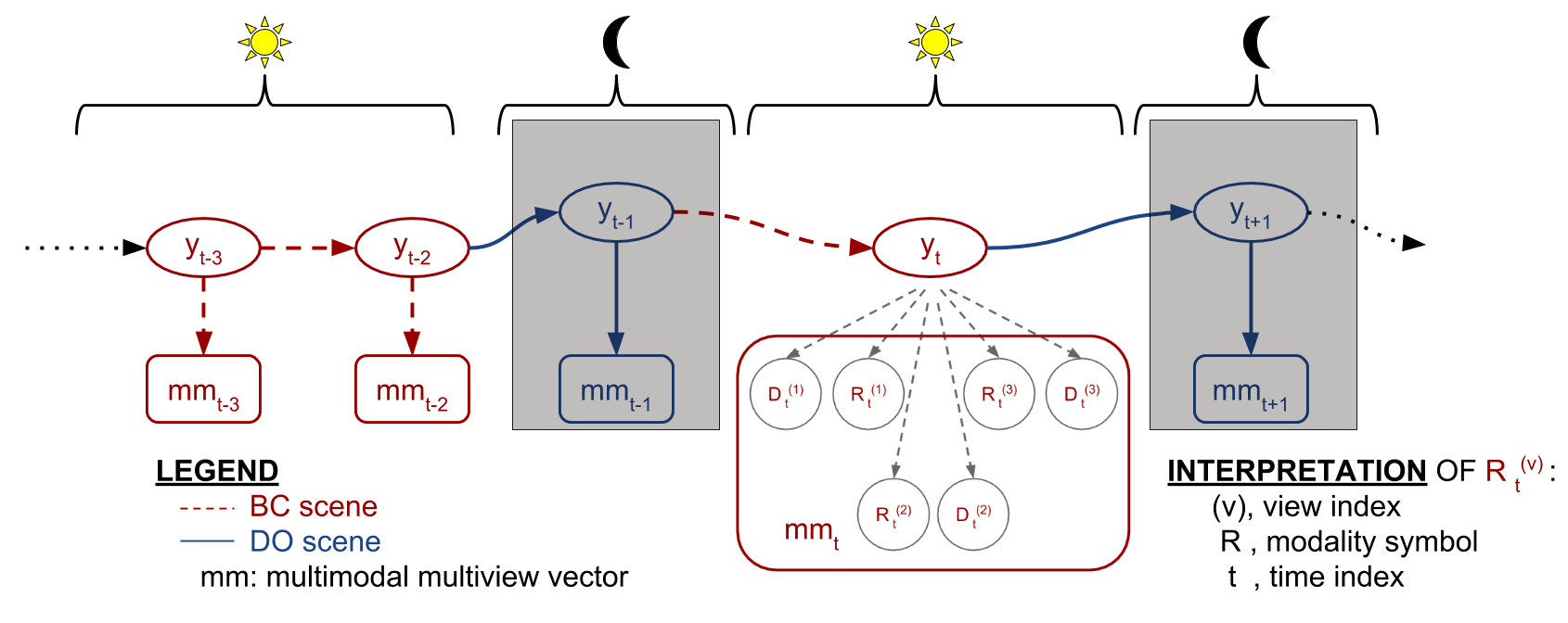}}
    \end{center}
    \vspace{-.5cm}
    \caption{Multimodal Multiview Hidden Markov Model (mmHMM) trellis. The variation in scene illumination between night and day are examples of scene changes.}
    \label{fig:mm_hmm_pgm}
\end{figure*}

\paragraph{Emission Probability Distribution $(\mathbf{B})$.}
The scene information is encoded into the emission probabilities. This information server to model moving from one scene condition to the next shown in Figure \ref{fig:mm_hmm_pgm}. The trellis shows two scenes, which  doubles the number of hidden states. The alternating blue and red lines (or solid and dashed lines) indicate transitions from one scene to the next. 

One limitation of HMMs is their lack of flexibility to model pose and transition (pseudo-poses) duration. Given an HMM in a known pose or pseudo-pose, the probability that it stays in there for $d$ time slices is: $P_i(d) = {(a_{ii})}^{d-1} (1-a_{ii})$, where $P_i(d)$ is the discrete probability density function (PDF) of duration $d$ in pose $i$ and $a_{ii}$ is the self-transition probability of pose $i$ \cite{rabiner1989tutorial}.


\subsection{Hidden Semi-Markov Models (HSMMs)} \label{sec:hsmm}
\begin{figure*}[t]
    \begin{center}
     	\fbox{\includegraphics[width=.8\linewidth]{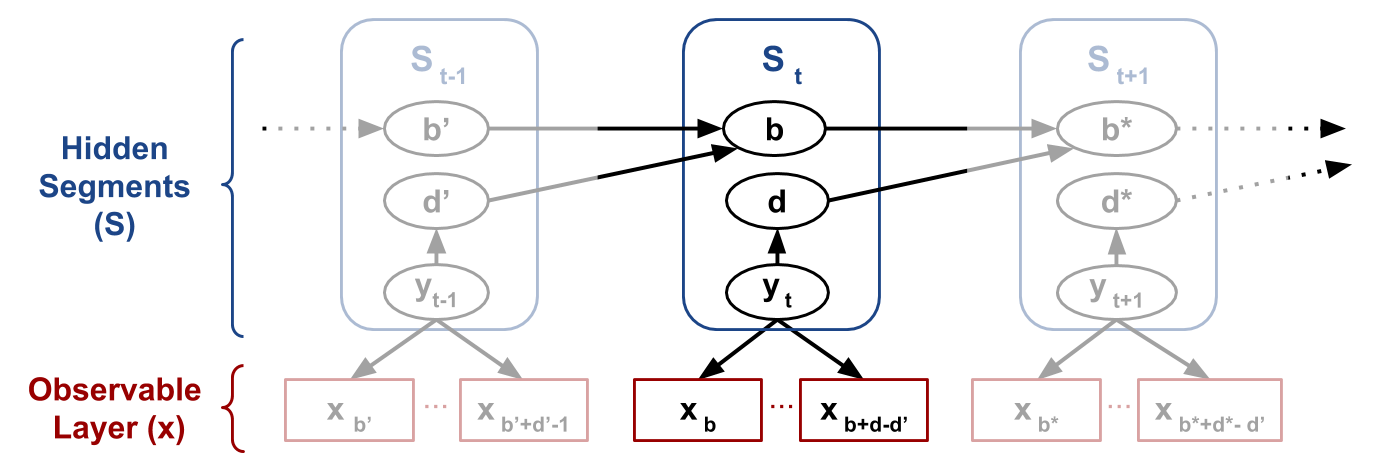}}
    \end{center}
    \vspace{-.5cm}
    \caption{HSMM diagram indicating the hidden segments $S_j$ indexed by $j$ and their elements $\{b_j, d_j, y_j\}$. The variable $b$ is the first detection in a sequence, $y$ is the hidden layer, $(x)$ is the observable layer containing samples from time $b$ to $b + d-d'$. The variables $b$ and $d$ are the observation's detection (time tick) and duration.}
    \label{fig:hsmm_pgm}
\end{figure*}

HSMMs are derived from conventiaonal HMMs to provide state duration flexbility. HSMMs represent hidden variables as segments, which have useful properties. Figure \ref{fig:hsmm_pgm} shows the structure of the HSMM and its main components. The sequence of states $y_{1:T}$ is represented by the segments $(S)$. A segment is a sequence of unique, sequentially repeated symbols. The segments contain information to identify when an observation is first detected and its duration based on the number of observed samples. The elements of the $j$-th segment $(S_j)$ are the indexes (from the original sequence) where the observation ($b_j$) is detected, the number of sequential observations of the same symbol ($d_j$), and the state or pose ($y_j$). For example, the sequence $y_{1:8} = \{ 1,1,1,2,2,1,2,2\}$ is represented by the set of segments $S_{1:U}$ with elements $S_{1:J}=\{S_1, S_2, S_3, S_4\} = \{(1, 3, 1), ~(4, 2, 2), ~(6, 1, 1), ~(7, 2, 2)\}$. The letter $J$ is the total number of segments and the total number of state changes. The elements of the segment $S_1=(1,3,1)$ are, from left to right: the index of the start of the segment (from the sequence: $y_{1:8}$); the number of times the state is observed; and the symbol.

\subsubsection{HSMM elements}
The hidden variables are the segments $S_{1:U}$, the observable variables are the features $X_{1:T}$. Their joint probability is:

\begin{equation} \label{eqn:hsmm}
    \begin{split}
        P\big(S_{1:U},X_{1:T}\big) = & P\big(Y_{1:U}, b_{1:U}, d_{1:U}, X_{1:T}\big)                              \\
        P\big(S_{1:U},X_{1:T}\big) = & P(y_1) P(b_1) P(d_1|y_1) \prod\limits_{t=b_1}^{b_1 + d_1 +1} P(x_t | y_1) *\\
                                     & \prod\limits_{u=2}^{U} P(y_u | y_{u-1}) P\big(b_u|b_{u-1}, d_{u-1}\big)   *\\ 
                                     & P\big(d_u|y_u\big) \prod\limits_{t=b_u}^{b_1 + d_1 +1} P(x_t | y_u), 
    \end{split}
\end{equation}

\noindent where $U$ is the sequence of segments such that $S_{1:U} = \{S_1, S_2, ..., S_U\}$ for $S_j = \big(b_j, d_j, y_j\big)$ and with $b_j$ as the start position (a bookkeeping variable to track the starting point of a segment), $d_j$ is the duration, and $y_j$ is the hidden state ($\in \{1, ..., Q\}$). The range of time slices starting at $b_j$ and ending at $b_j + d_j$ (exclusively) have state label $y_j$. All segments have a positive duration and completely cover the time-span $1:T$ without overlap. Therefore, the constraints $b_1 = 1$, $\sum\limits_{u=1}^U$ and $b_{j+1}=b_j+d_j$ hold. 

The transition probability $P(y_u|y_{u-1})$, represents the probability of going from one segment to the next via:

\begin{equation}
    \mathbf{A}: P\big(y_u=j | y_{t-u}=i\big) \equiv a_{ij}
\end{equation}

The first segment ($b_u$) always starts at 1 ($u=1$). Consecutive points are calculated deterministically from the previous point via:

\begin{equation} \label{eqn:bu1}
        P\big(b_u=m|b_{u-1} = n, d_{u-1}=l\big) = \delta \big(m,n+l\big)
\end{equation}

\noindent where $\delta(i,j)$ is the Kroenecker delta function (1, for $i=j$ and 0, else).

The duration probability is now given by $P (d_u=l | y_u = i) = P_i(l)$. DECU uses $P_i(l) = \mathcal{N}(\mu,\sigma)$.

\subsubsection{Parameter Learning} Learning is based on maximum likelihood estimation (mle). The training sequence of key frames is fully annotated, including the exact start and end frames for each segment $X_{1:T}, Y_{1:T}$. To find the parameters that maximize $P\big(Y_{1:T}, X_{1:T} | \theta \big)$, one maxizes the likelihood parameters of each of the factors in the joint probability. In particular, the observation probability $P\big(x^n | y=i\big)$, is a Bernoulli distribution whose max likelihood is estimated via: 

\begin{equation} \label{eqn:mu}
    \mu_{n,i} = \frac{\sum_{t=1}^{T} x_{t}^{i} \delta\big(y_t,i \big)} {\sum_{t=1}^{T} \delta\big(y_t,i \big)},
\end{equation}

 \noindent where $T$ is the number of data points, $\delta(i,j)$ is the Kroenecker delta function, and $P\big(y_t=j | y_{t-1}=i\big)$ is the multinomial distribution with:

\begin{equation}\label{eqn:aij}
 a_{ij} = \frac{\sum_{n=2}^{N} \delta\big( y_n,j\big) \delta\big(y_{n-1},i \big)} {\sum_{n=2}^{N} \delta\big(y_{t-1},j \big)}
\end{equation}

\subsubsection{HSMM Inference} \label{sec:hsmm}

\paragraph{HSMM Viterbi}
The segment notation is used to represent state sequences. The inference objective is to find the state sequence that maximizes $P\big( S_{1:U}, X_{1:T} | \theta \big)$.
The duration is not known for a new sequence of observations. The sequence corresponding to the duration with the highest probability is determined at each time step by iterating over all possible durations from 1 to a prefix duration $D$. This information is stored as follows:

\begin{equation}\label{eqn:tau_max}
 \tau_{t,d} = \underset{ s_1, ... , s_{k-1} } {\max} P \Big( X_{1:t}, s_{1:k} = \big( t-d+1, d,i\big) | \theta \Big), 
\end{equation}

\noindent which represents the highest probability of a sequence of $K$ segments, where the final segment started at $t-d+1$, has duration $d$ and label $i$. \\

\noindent \textbf{NOTE:} just as with conventional HMMs, it is sufficient to only keep track of the max probability of ending in state $s_{k-1}$  to effectively compute the max probability of ending up in state $s_k$.\\

The state label (for a pose or pseudo-pose) of the previous segment is stored in array $\zeta_t (d,i)$. The max probability duration ($\delta$) is computed via: 
\begin{equation}\label{eqn:duration_prob}
 \delta_{t}(i) = \underset{ s_1, ... , s_{k-1} } {\max} P \Big( x_{1:t}, s_{1:k} = \big( t-d^* +1, d^*,i\big) | \theta \Big), 
\end{equation}

\noindent where $d^*$ is the duration with the highest probability at time $t$ for state $i$. The best duration is stored in $\phi_t(i)$ and the label of the previous segment is stored in $\psi_{t} (i)$.

\subsubsection{Finding the Best Sequence}
The complete procedure for finding the best sequence is described in the following procedure:

\paragraph{Initialization:} The probability of label of the first segment is given by the initial state distribution $\pi$.
\begin{equation*}\label{eqn:tau_ini}
        \tau_{t,d} = \pi_{i} P_i(d) \prod_{t=1}^{T}P\big(x_t | y_t\big)
\end{equation*}

\begin{equation*}\label{eqn:zeta_ini}
    \zeta_d(d,i) = 0
\end{equation*}

\paragraph{Recursion:} Iterate over all possible durations at each step.
\begin{equation*}\label{eqn:tau_rec}
        \tau_{t,d} = \underset{1 \leq i \leq Q } {\max}  \big[ \delta_{t-d}(i) a_{ij} \big] P_{j}(d) \prod_{m=m_1}^{t}P\big( \vec{x}_m | y_m = j\big),
\end{equation*}

\noindent where $m_1 = t-d+1$ and 

\begin{equation*}\label{eqn:zeta_rec}
    \zeta_d(d,i) = \underset{1 \leq i \leq Q } {\arg \max}  \big[ \delta_{t-d}(i) a_{ij} \big]
\end{equation*}

\noindent The duration with the highest probability is estimated via:
\begin{equation*}\label{eqn:delta_rec}
    \delta_t(i) = \underset{1 \leq d \leq D } {\max}  \big[ \delta_{t-d}(i) a_{ij} \big],
\end{equation*}
\noindent which represents the best segment. The variable $d^*$ is the duration with the highest probability at time $t$ for state $i$. The best duration for state $i$ at time $t$ is estimated via:
\begin{equation*}\label{eqn:phi_rec}
    \phi_t(i) = \underset{1 \leq d \leq D } {\arg \max} ~\tau_{d,t}(i).
\end{equation*}

\noindent Finally, $\psi_t(i) = \zeta_t\big(\phi_t(i),i\big)$ is the label corresponding to the best duration for time $t$ and state $i$.
 
\paragraph{Termination:} Estimate the state with the highest probability in the last timeslice.
\begin{equation*}\label{eqn:terminate}
    \begin{split}
        P^* =   & \underset{1 \leq i \leq Q } {\max} [ \delta_T(i) ] \\
        y^*_T = & \arg \max [ \delta_T(i) ] \\
        t     = & T \\
        u     = & 0\\
    \end{split}
\end{equation*}

\paragraph{Backtracking:} Starting from termination look up the duration and previous states stored in variables $\phi$ and $\psi$.

\begin{equation*}\label{eqn:backtrack}
    \begin{split}
        d^*_t = & \phi_t \big( y^*_t \big) ] \\
        s^*_u = & \big( t-d^*_t + 1, d^*_t, y^*_t \big) \\
        t     = & t - d^*_t \\
        u     = & u-1\\
        y^*_t = & \phi_{t+d}(y^*_{t+d})
    \end{split}
\end{equation*}

Negative indexing is used for the segments because the number of segments is not known in advance. However, this is corrected after inference by adding $|S^*|$ to all indices.

\subsection{Key Frame ($KF$) Selection.}
\begin{figure*}[t]
    \begin{center}
     	\fbox{\includegraphics[width=.85\linewidth]{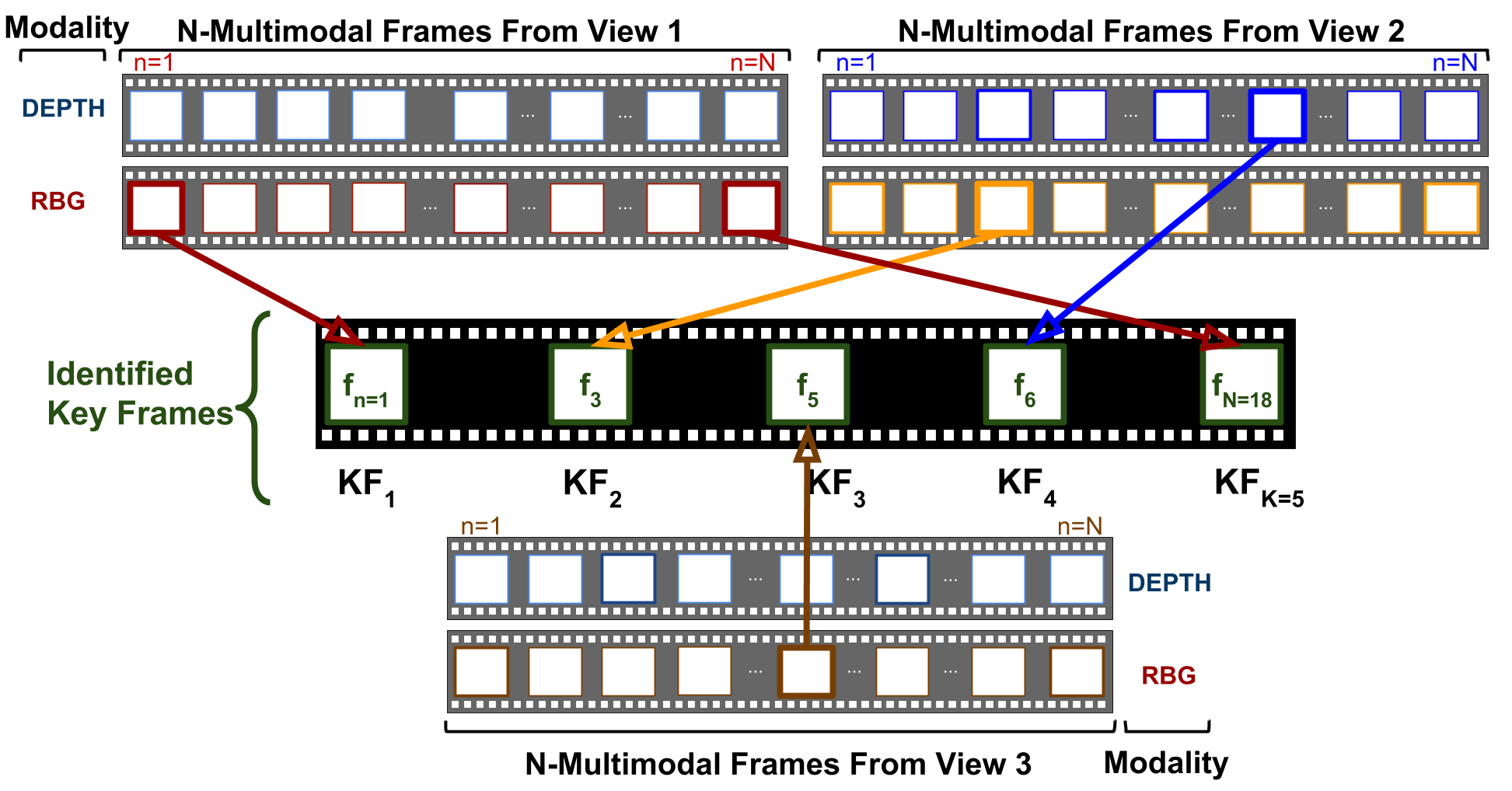}}
    \end{center}
    \vspace{-.5cm}
    \caption{Selection key frames for the represention of transitions between two poses. The key frame selection is based on Algorithm \ref{algo:keyframes}. This figure show an example of how the algorithm is used to identify five key frames from three views and two modalities. In this example, the first two key frames are extracted from the RGB video from the first camera (View 1). Subsequent key frames are selected from the depth video from the second camera (View 2) and from the RGB video from the third camera (View 3).}
    \label{fig:keyframes}
\end{figure*}

Data collected from pose transition is very large and often repetitive, since the motion is relatively slow and subtle. The pre-processing stage incorporates a key frame estimation step that integrates multimodal and multiview data. The algorithm used to select a set ($KF$) of $K$-transitory frames is shown in Figure \ref{fig:keyframes} and detailed in Algorithm \ref{algo:keyframes}. The size of the key frame set is determined experimentally ($K=5$) on the feature scape using Inception vectors.

Let $\mathcal{X} = \{x^{(v)}_{m,n} \}_f$ be the set of training features extracted from $V$ views and $M$ modalities over $N$ frames and let $P_i$ and $P_o$ represent the initial and final poses. The transition frames are indexed by $n$, $1 \leq n \leq |N|$; views are indexed by $v$, $1\leq v \leq |V|$ and modalities are indexed by $m$, $ 1 \leq m \leq |\mathcal{M}|$. Algorithm \ref{algo:keyframes} uses this information to identify key frames. Experimental evaluation of $|KF|$ size is shown in Figure \ref{fig:keyframe_performance}. Key frames are the most informative and discriminant frames across all views and modalities.

\begin{algorithm}
{\small
\SetAlgoLined
\textbf{Input:} $\mathcal{X}$, set of mm features and dissimilarity threshold $th$\;
\KwResult{ $KF = \{\text{Key Frames}\}_K$, $K \geq 1$  }
\textbf{Initialize:} $KF = \{\text{empty}\}_K$, $K \geq 1$ and $count = 0$ \; 

\textbf{Stage 1:} Modality ($m$) and View ($v$) Selection\;
    \For{$ 1 < v < V $ and $ 1< m < M$}{
            $D^{(v)}_{m} =$ euclid$(x^{(v)}_{mn_i}, x^{(v)}_{mn_o})$, $n_i = 1, n_o = N$\; 
            }
    
    $\hat{v}$, $\hat{m} = \text{ max } D^{(v)}_{m} > th$\;
    
    $\{x^{(\hat{v})}_{\hat{m}n_1}, x^{(\hat{v})}_{\hat{m}n_N}\} \rightarrow FK$ \;
    
\textbf{Stage 2:} Find Complementary Frames to $KF$ \;
\For{ $1 < v < V $ and $1 < m < M$ and $1 < n < N $}{
    $D_1 = D^{(v)}_{m,n_1} =$ euclid$(x^{(v)}_{mn_1}, x^{(v)}_{mn})$\;
    $D_2 = D^{(v)}_{m,n_N} =$ euclid$(x^{(v)}_{mn_N}, x^{(v)}_{mn})$\;
    }

Sort $D_1 = \{d_1 > d_2 >... > d_{N-2}\}$ descending\;
Sort $D_2 = \{d_1 > d_2 >... > d_{N-2}\}$ descending\;

$d_i \rightarrow KF$ if $\frac{d_i}{d_j} > th$, \text{ for }  $1 < i,j < N-2$ \;

\textbf{Stage 3:} Find Center Frame (i.e., Motion Peak)\;
\For{$KF_2$ and $KF_{K-1}$}{
    Use Stage 2 to compute $D_3$ and $D_4$\;
    \If{$\text{max} (D_3, D_4) > 0$)}
        {$\text{max }(D_3, D_4) \rightarrow KF$\;}}
}
\caption{Multimodal multiview key frame selection using euclidean dissimilarity measure. The algorithm is applied at training with labeled frames to estimate the number and indexes of key frames across views and modalities.} \label{algo:keyframes}
\end{algorithm}

\section{Experimental Results and Analysis}
Experiments are performed to validate the feature selection, the keyframe set size (i.e., number of states) representing a transition, and the summarization performance of DECU in a real and a mock-up ICU environment. 
\subsection{Static Pose Analysis - Feature Validation} \label{sec:exp_FeatValidation}
Static sleep-pose analysis is used to compare the DECU method to previous studies. Couple-Constrained Least-Squares (cc-LS) and DECU are tested on the dataset from \cite{torres2016EyeCU}. Combining the cc-LS method with deep features extracted from two common network architectures improved classification performance over the HOG and gMOM features in dark and occluded (DO) scenes by an average of eight percent with Inception and four percent with Vgg. Deep features matched the performance of cc-LS (with HOG and gMOM) for a bright and clear (BC) scenario shown in Table \ref{table:feat_validation}.

\begin{table}[t]
{\small %
\centering
\begin{tabular}{|C{1.5cm}||C{1.5cm}|C{1.5cm}|C{1.5cm}|} \hline 

\hline  
\multicolumn{4}{c}{\bf{Feature Suitability Evaluation with cc-LS \cite{torres2016EyeCU}}}         \\ \hline\hline
Scene   & HOG + gMOM  & Vgg           & Inception                            \\ \hline
BC      & 100         & 100           & 100                                  \\
DO      & 65          & 69 (+4)       & 73 (+8)                              \\ 
\hline
\end{tabular}
\caption{Evaluation of deep features for sleep-pose recognition tasks using the cc-LS method from \cite{torres2016EyeCU} in dark and occluded (DO) scenes using. The performance of HOG and gMOM is compared to the performance of the Vgg and Inception features.}
\label{table:feat_validation}
}
\end{table}

\subsection{Key Frame Performance}\label{exp:keyframes}
\begin{figure}[t]
    \begin{center}
     	\fbox{\includegraphics[width=.95\linewidth]{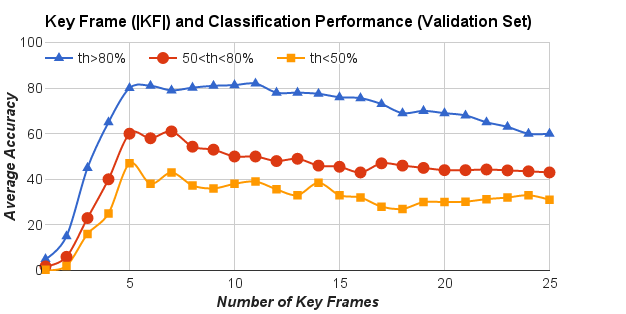}}
    \end{center}
    \vspace{-.5cm}
    \caption{Performance of the DECU framework for the fine motion summarization based on the number of key frames used to represent transitions and rotations between poses.}
    \label{fig:keyframe_performance}
\end{figure}

The key frame set size ($|KF|=5$) and key frame dissimilarity threshold ($th \geq .8$) affects DECU performance. Figure \ref{fig:keyframe_performance} shows the effect of these parameters.

\subsection{Summarization Performance}
The mock-up ICU allows staging the motion and scene condition variations without disturbing patients in the medical ICU. A sample test sequence is shown in Figure \ref{fig:ICU}(c) and summarization history results are shown in Figure \ref{fig:summarization_history} for (a) the mock-up and (b) the real ICU environments. The pose numerical symbols are shown in Table \ref{table:pose_symbols}

\begin{table}
{\small %
\centering
\begin{tabular}{|C{1.5cm}||C{3cm}|} \hline 
\multicolumn{2}{c}{\bf{DECU: Pose History Summarization}}       \\ \hline\hline
Symbol         & Pose Name                         \\ \hline
   0           & Aspiration                        \\
+1 / -1        & Soldier (+Up / -Down)             \\ 
+2 / -2        & Yearner (+R / -L)                 \\ 
+3 / -3        & Log (+R, -L)                      \\ 
+4 / -4        & Faller (+Up / -Down)              \\ 
+5 / -5        & Other / Background                \\    
+6 / -6        & Fetal (+R / -L)                   \\

\hline
\end{tabular}
\caption{Pose symbols and descriptions used for ICU pose history summarization.}
\label{table:pose_symbols}
} 
\end{table}

\subsection{Summarization History}
\begin{figure}[!h]
    \begin{center}
     	\fbox{\includegraphics[width=.9\linewidth]{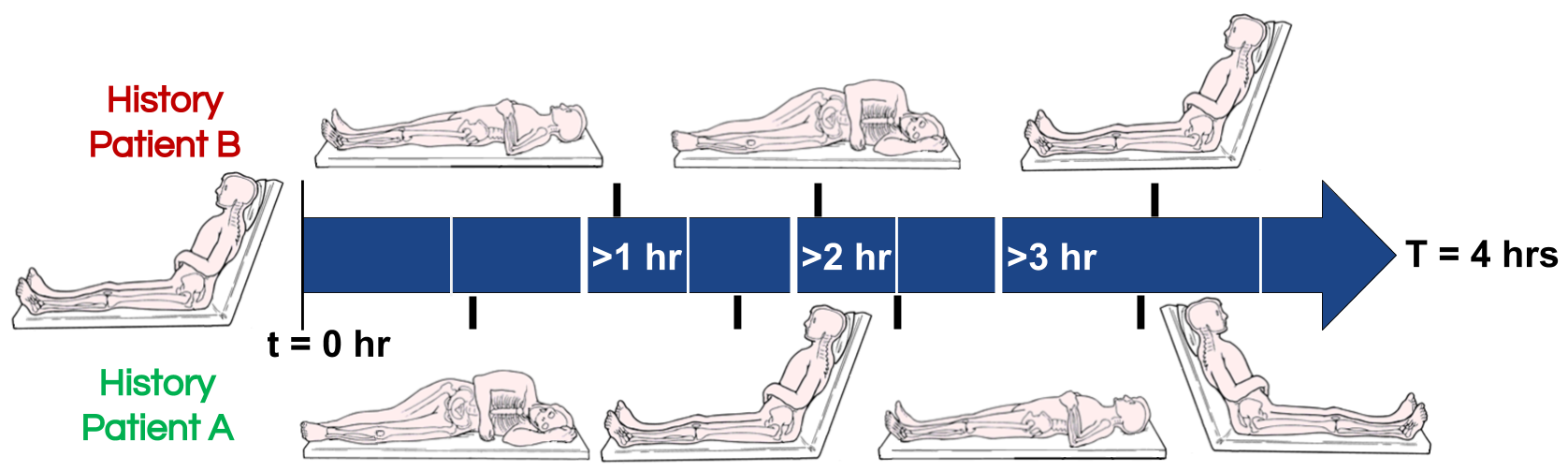}}
    \end{center}
    \vspace{-.5cm}
    \caption{Pose history summarization log for patient motion analysis in medical ICUs. }
    \label{fig:ICU_summary_illustration}
\end{figure}

History summarization is the coarser time resolution and its overall objective is shown in Figure~\ref{fig:ICU_summary_illustration}.

\begin{figure*}[hbtp]
    \begin{center}
        \begin{tabular}{c}
             \fbox{\includegraphics[width=.775\paperwidth]{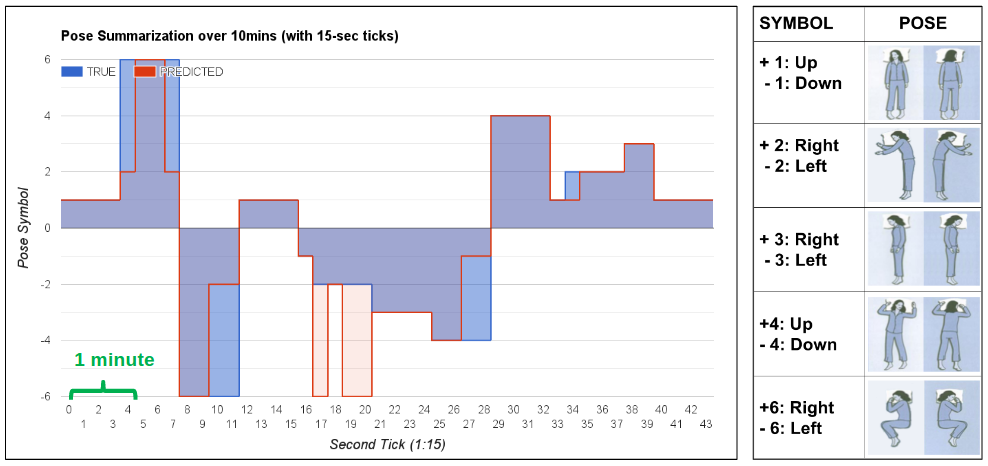}} \\
             (a) \\
            \fbox{\includegraphics[width=.775\paperwidth]{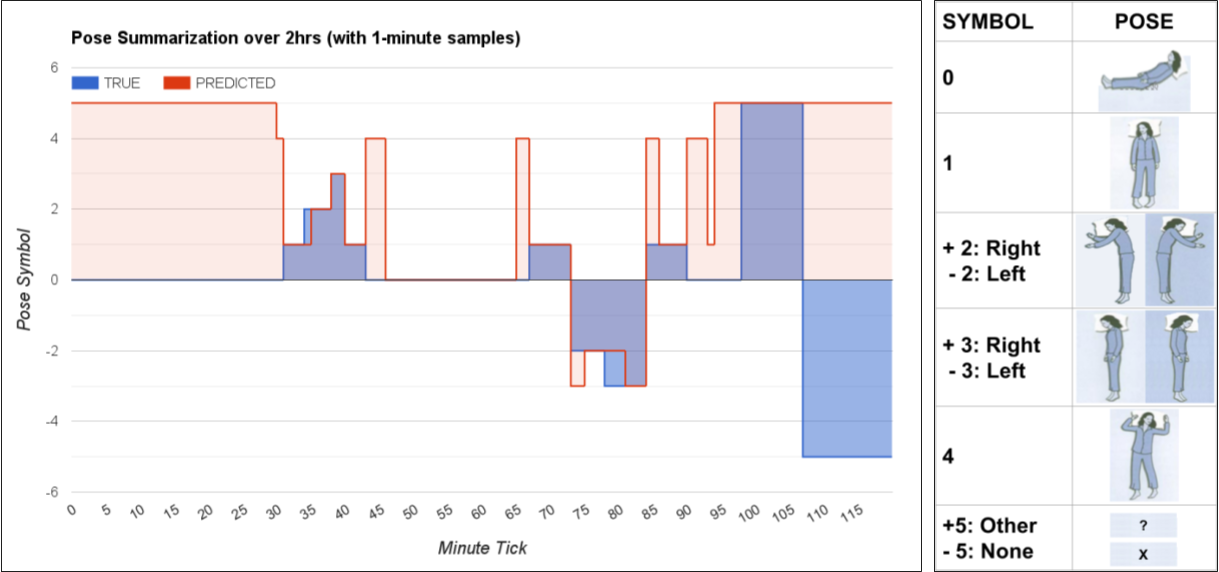}} \\
             (b) \\
        \end{tabular}
    \end{center}
    \caption{Performance of DECU pose history summarization in a the mock-up ICU with bright and clear conditions over a 10-minute time-span (a) and in the real ICU using multimodal data under natural scene conditions over a two-hr time-span (b). Note that the set of patient poses is reduced for the real ICU and the summarization performance is limited to a maximum session of two hours to avoid disrupting the Braden-scale protocol.}
    \label{fig:summarization_history}
\end{figure*}

\paragraph{Pose History Summarization in the Mock-Up ICU.}
This summarization requires two parameters: sampling rate and pose duration. The experiments are executed with a sampling rate of one second and pose duration of 10 seconds. A pose is assigned a label if consistently detect 80 percent of the time, else the assigned label is "other". Poses not consistently detected are ignored (low confidence). The mock-up experiment uses a randomly selected scene and sequence of poses, which can range from two to ten poses. The pose duration is also set at random  and includes one scene transition (BC to DO or DO to BC). A sample (long) sequence is shown in Figure \ref{fig:ICU} (c) and its history summarization performance is shown in Table \ref{table:history} and Figure~\ref{fig:summarization_history}(a).

\begin{table}[!h]
\centering
{\small %
\begin{tabular}{|C{1cm}||C{3cm}|} \hline 
\multicolumn{2}{c}{\bf{DECU: Pose History Summarization}}       \\ \hline\hline
Scene   & Average Detection Rate    \\ \hline
BC      & 85                        \\
DO      & 76                        \\ 
\hline
\end{tabular}
\caption{Pose history summarization performance (percent accuracy) of the DECU framework in bright and clear (BC) and dark and occluded (DO) scenes. The sequences are composed of 10 poses with duration ranging from 10 seconds to 1 minute. The sampling rate is one second.}
\label{table:history}
} 
\end{table}

\begin{figure*}[hbtp]
    \begin{center}
    
        \begin{tabular}{c}
            \fbox{\includegraphics[width=.975\linewidth]{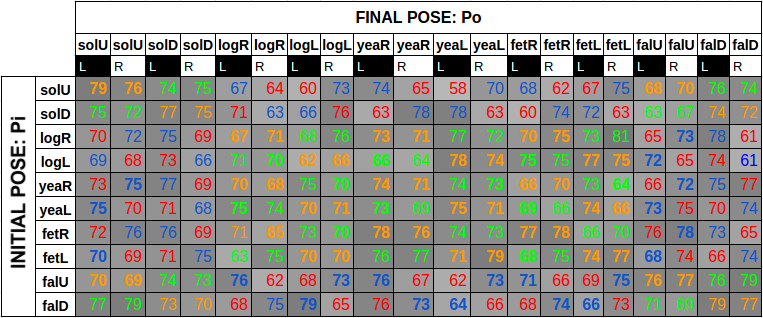} } \\
            (a) \\
            \fbox{\includegraphics[width=.975\linewidth]{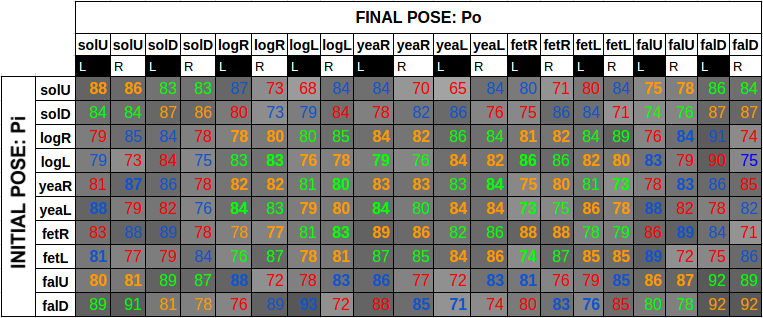} } \\
            (b) \\
            \fbox{\includegraphics[width=.975\linewidth]{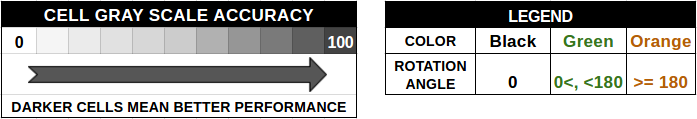} } \\ 
            (c) \\
        \end{tabular}
    \end{center}
    \caption{Performance of DECU in the mock-up ICU under bright and clear conditions. Detection results are obtained using (a) single view and (b) multiview data. The cells are gray scaled to indicate detection accuracy. The color coded scale and the legend are shown in (c). Note that overall detection rates increase with longer rotation angles and decrease when rotation motion requires the actors to face the bed (i.e., cameras record actor backs).}
    \label{fig:BC_results_transitions}
\end{figure*}

\begin{figure*}[hbtp]
    \begin{center}
    
        \begin{tabular}{c}
            \fbox{\includegraphics[width=.975\linewidth]{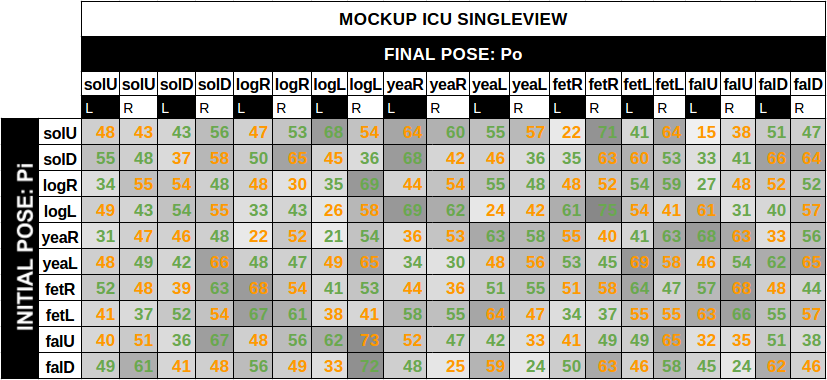} } \\
            (a) \\
            \fbox{\includegraphics[width=.975\linewidth]{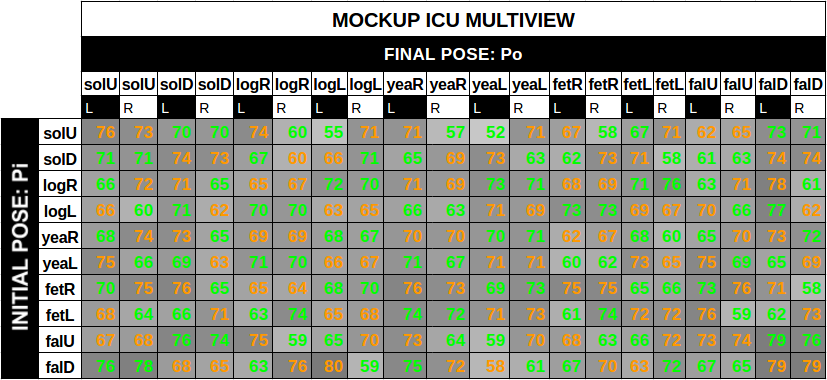} } \\
            (b) \\
            \fbox{\includegraphics[width=.975\linewidth]{figx_scale_and_legend.png} } \\ 
            (c) \\
        \end{tabular}
    \end{center}
    \caption{Performance of DECU in the mock-up ICU under dark and occluded conditions. Detection results are obtained using (a) single view and (b) multiview data. The cells are gray scaled to indicate detection accuracy. The color coded scale and the legend are shown in (c). Again, detection rates increase with longer rotation angles and decrease when rotation motion requires the actors to face the bed (i.e., cameras record actor backs).}
    \label{fig:DO_results_transitions}
\end{figure*}

\paragraph{Pose Transition Dynamics: Motion Direction.}
The analysis and pose transitions and rotation directions are important to physical therapy and recovery rate analysis. The performance of DECU summarizing fine motion to describe transitions between poses for a bright and clear scene and a dark and occluded scene are shown in Figs. \ref{fig:BC_results_transitions} and \ref{fig:DO_results_transitions}. Results for each of the figures are shown for (a) singleview and (b) multiview data. The bottom row (c) shows the gray scale and the color-font legend.

\paragraph{Summarization of Transitions in the real-ICU}
Note that it is logistically impossible to control ICU work flows and to account for unpredictable patient motion. ICU patients are not free to rotate, which reduces the set of pose transitions (unavailable transitions are marked N/A). The set of poses for the history summary require that a new pose be included (aspiration). Figure \ref{fig:ICU_summary_illustration} (b) shows the overall clinical objective behind the pose history summarization.

\begin{figure*}[!hbtp] 
    \begin{center}
        \begin{tabular}{c}
             \fbox{\includegraphics[width=.975\linewidth]{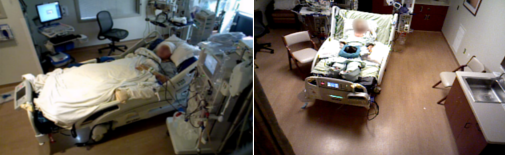}} \\
             (a) \\
            \fbox{\includegraphics[width=.975\linewidth]{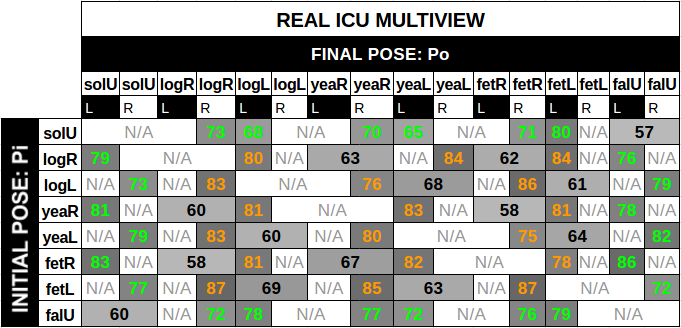}} \\ 
            (b) \\
            \fbox{\includegraphics[width=.975\linewidth]{figx_scale_and_legend.png}} \\
            (c) \\            
        \end{tabular}
    \end{center}
    \caption{Performance of DECU pose transition summarization in a real ICU shown in (a) using multimodal data under natural scene conditions. The detection scores are shown in (b), where the cells are gray scaled to indicate detection accuracy. The font color indicates rotation angle range and N/A indicated the pose is not available. The number of poses is reduced due to patient health conditions and inability to move. The grading color scale and font-color legend are shown in (c).}
    \label{fig:ICU_results_transitions}
\end{figure*}

The real medical ICU environment is shown in Figure \ref{fig:ICU_results_transitions} (a). DECU's fine motion summarization results for two patients are shown in Figure \ref{fig:ICU_results_transitions}(b) and the quantified detection accuracies are shown in Figure \ref{fig:ICU_results_transitions} (c). The blue trace represents the true transition labels and the red trace indicates the predicted labels. Table \ref{table:pose_symbols} has the pose symbols and descriptions used in the summarization plot.

\section{Conclusion}\label{sec:Conclusion}
\vspace{-.25cm}
This work introduced the DECU framework to analyze patient poses in natural healthcare environments at two motion resolutions. Extensive experiments and evaluation of the framework indicate that the detection and quantification of pose dynamics is possible. The DECU system and monitoring algorithms are currently being tested in real ICU environments. The performance results presented in this study support its potential applications and benefits to healthcare analytics. The system is non-disruptive and non-intrusive. It is robust to variations in illumination, view, orientation, and partial occlusions. DECU is non-obtrusive and non-intrusive but not without a cost. The cost is noticed in the most challenging scenario where a blanket and poor illumination block sensor measurements. The performance of DECU to monitor pose transitions in dark and occluded environments is far from perfect; however, most medical applications that analyze motion transitions, such as physical therapy sessions, are carried under less severe conditions.
\vspace{-.25cm}
\paragraph{Future Work}\label{sec:FutWork}
Future studies will investigate the recognition and analysis patient motion in similar challenging scenarios using recurrent neural networks, incorporate additional modalities, and integrate natural language understanding to analyze ICU events.

\clearpage

\clearpage

\clearpage
\paragraph{Acknowledgements.} Research is funded in part by the Army Research Laboratory under Cooperative Agreement Number W911NF-09-2-0053 (the ARL Network Science CTA). The views and conclusions contained in this document are those of the authors and should not be interpreted as representing the official policies, either expressed or implied, of the Army Research Laboratory or the U.S. Government. The U.S. Government is authorized to reproduce and distribute reprints for Government purposes notwithstanding any copyright notation here on. The authors want to thank Richard Beswick, PhD (Director of Research), Paula Gallucci (Medical ICU Nurse Manager), Mark Mullenary (Director Biomedical Engineering), and Leilani Price, PhD (IRB Administrator) from Santa Barbara Cottage Hospital for their help and patience identifying and recruiting patients and ensuring HIPAA compliance.

\bibliographystyle{ieee}
{\small

}
\end{document}